\documentclass[letterpaper]{article}

\usepackage{aaai2027}

\usepackage[hyphens]{url}
\usepackage{graphicx}
\usepackage{natbib}
\usepackage{caption}

\usepackage{algorithm}
\usepackage{algorithmic}

\usepackage{newfloat}
\usepackage{listings}

\DeclareCaptionStyle{ruled}{
    labelfont=normalfont,
    labelsep=colon,
    strut=off
}

\floatstyle{ruled}
\newfloat{listing}{tb}{lst}{}
\floatname{listing}{Listing}

\usepackage{booktabs}

\usepackage{amsmath}
\usepackage{amssymb}
\usepackage{mathtools}
\usepackage{amsthm}
\usepackage{amsfonts}

\usepackage{multirow}
\usepackage{wasysym}

\usepackage{xcolor}

\usepackage{float}
\usepackage{dblfloatfix}

\usepackage{amsmath}
\usepackage{amssymb}
\usepackage{mathtools}
\usepackage{amsthm}
\usepackage{multirow}
\usepackage{wasysym}
\usepackage{float}
\usepackage{xcolor}
\usepackage{amsfonts}
\usepackage{amsmath}
\usepackage{multirow}
\usepackage{booktabs}
\usepackage{times}
\usepackage{helvet}
\usepackage{courier}
\usepackage{xcolor}
\usepackage{graphicx}
\usepackage[table]{xcolor}
\usepackage{caption}
\usepackage{graphicx}
\usepackage{dblfloatfix}

\title{
MixDiffusion: Mixing Diffusion-based Uni-condition Text-to-Image Generation Models for Multi-condition Image Synthesis
}

\author{
Pengcheng Wan,
Liang Han,
Lin Xu,
Bowen Xiao,
Liqiang Nie
}

\affiliations{
Harbin Institute of Technology (Shenzhen)\\
}

\begin{document}

	\twocolumn[{
		\renewcommand\twocolumn[1][]{#1}
		\maketitle
		\begin{center}
			\centering
			\includegraphics[width=0.8\linewidth]{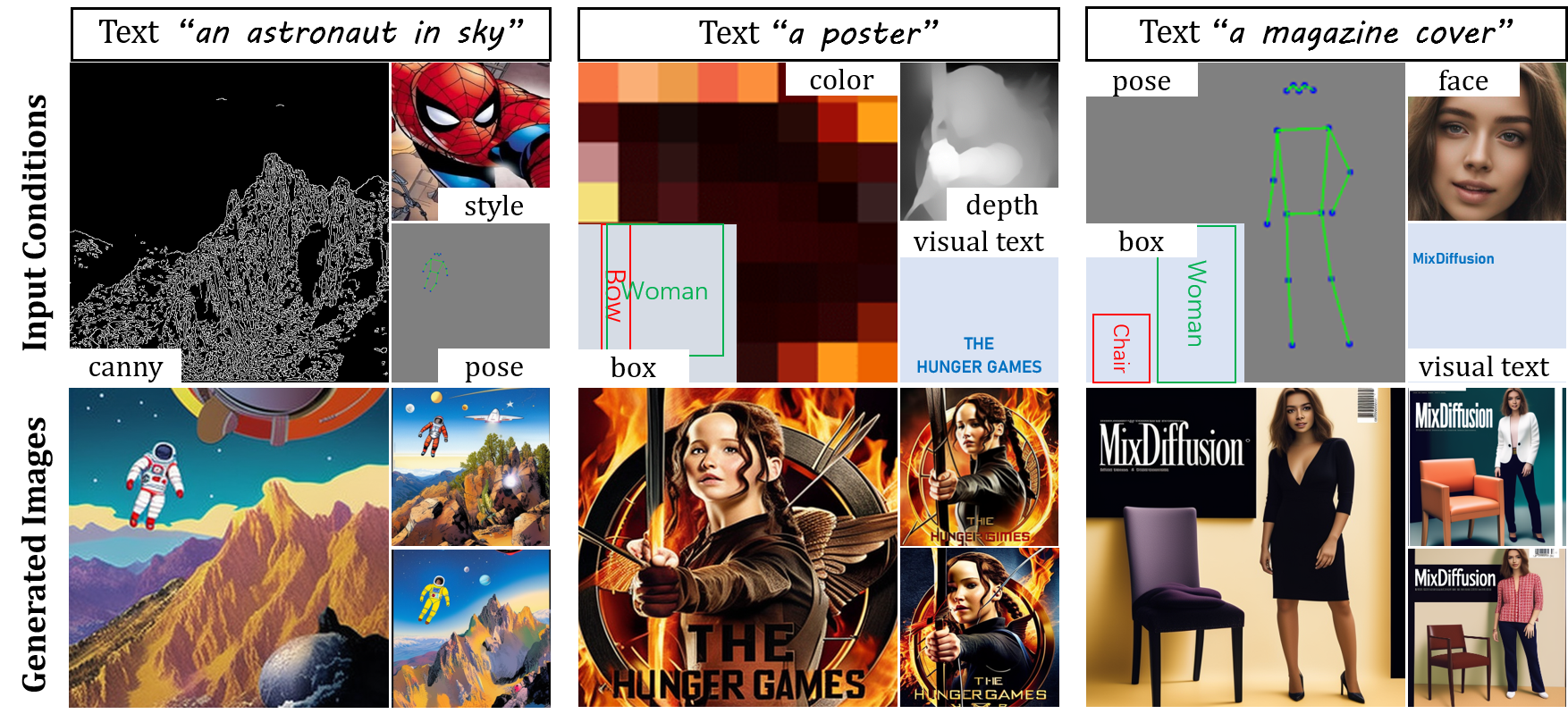}
			\captionof{figure}{
            \textbf{Images generated with our proposed MixDiffusion.} Each column shows a controllable image generation example. The upper part of each column shows the provided control conditions, and the lower part presents the generated images by the proposed MixDiffusion under these control conditions. MixDiffusion is training-free, and can support an unlimited number of control condition inputs theoretically.
            }
			\label{headfigure}
		\end{center}
	}]

\begin{abstract}

Recent advances in text-to-image (T2I) generation have enabled controllable image synthesis by incorporating conditions beyond text. However, most existing diffusion-based methods are limited to a single type of control condition (e.g., bounding boxes or keypoints), which restricts their flexibility. To address this limitation, we propose MixDiffusion, a training-free diffusion framework for multi-condition T2I generation. MixDiffusion theoretically supports an arbitrary number of control conditions—including bounding boxes, keypoints, sketches, depth maps, reference images, and text—by collaboratively integrating multiple pre-trained uni-condition diffusion models. The key insight of the proposed approach is to derive the predicted noise distribution in each denoising step of the diffusion-based multi-condition image generation model from the predicted noise distributions of multiple diffusion-based uni-condition models with a derived integration formula, which is supported by rigorous theory proof. Owing to its training-free nature, MixDiffusion is easy to deploy and readily extensible to new control modalities. 

\end{abstract}

\section{Introduction}
How can one accurately convey the images in his mind to another person? As the saying goes, ``a picture is worth a thousand words''. Drawing is undeniably a more intuitive and precise way than textual description. Thanks to the significant advancements in the T2I synthesis community \cite{GAN,DDPM,DDIM}, it is now possible to generate realistic and high-fidelity images within seconds using advanced models. For example, T2I synthesis models such as Stable Diffusion \cite{sd,sdxl}, Imagen \cite{Imagen}, DALL-E \cite{Dall-E2,Dall-E3} and FLUX \cite{flux2024,flux-2-2025} have shown remarkable capabilities in creating high-quality images with impressive aesthetic and realistic features. However, these models often struggle with generating images with controllable details, e.g., generating particular visual text in the image, placing objects in precise locations, or rendering people in specific poses.

In response to these limitations, substantial progress has been made in controllable image generation. By providing diffusion models with additional inputs beyond textual descriptions, e.g., bounding boxes in BoxDiff \cite{BoxDiff} and InstanceDiffusion \cite{InstanceDiffusion}, keypoints in the GFLA framework \cite{posemodel1} and PIDM \cite{posemodel2}, and segmentation masks in ALDM \cite{ALDM} and SceneComposer \cite{SceneComposer}, the image generation models have been able to generate images following specific control condition inputs. Since these models support only one customized control condition besides textual description to guide the generation, we name them as ‘uni-condition image generation models’ in this paper.

Despite achieving notable success through either fine-tuning or training from scratch, one of the biggest limitations of the uni-condition image generation models is that only one type of condition is supported to guide the image generation. For example, if we aim to generate a fashion magazine cover featuring a charming girl with particular pose using DreamPose \cite{DreamPose}, which accepts keypoints as an input control condition, the resulting image would likely satisfy the desired pose. However, it is very challenging or even impossible for DreamPose to generate a girl with a specific face (e.g., a face in a reference image) if the generated girl is not a celebrity. Recently, a few diffusion-based image generation models, e.g., Uni-ControlNet \cite{zhao2024uni} and Diffblender \cite{diffblender}, are proposed, which can support more than one control conditions besides text description. Unfortunately, these models can only accept a very limited number of control conditions, and the control conditions are fixed for each model. Retraining the model to accommodate additional control conditions is both time-consuming and expensive. Moreover, as user demands evolve, the range of possible conditions for controllable image generation expands, resulting in countless potential combinations. 

To overcome the above limitations of the existing controllable image generation models, we introduce MixDiffusion, a training-free framework that supports an unlimited number of control condition inputs by mixing multiple pre-trained diffusion based uni-condition image generation models to collaborate in generating a single image. Moreover, we dive deep into the diffusion model theory and design an integration strategy by mathematically deriving an integration formula for the proposed MixDiffusion, which is more reasonable and theoretically interpretable, and boosts MixDiffusion to achieve superior generation results when combining multiple models. As shown in Fig.~\ref{headfigure}, the images generated by MixDiffusion perfectly satisfy all control condition inputs.

The main contributions of this work are as follows:
\begin{itemize}
\item  A training-free framework is proposed for integrating multiple diffusion based uni-condition T2I generation models, supporting flexible and an unlimited number of control conditions to intricately control the image generation. This fills the gap between real-world applications requiring multi-condition controllable image generation and the current T2I generation models that predominantly support only one or two control conditions. Additionally, the training-free property of the proposed framework is highly convenient for deployment and for adding new type of control conditions as needed. 
\item We extend the theory system of diffusion models by deriving the predicted noise distribution of a diffusion based multi-condition image generation model from the predicted noise distributions of multiple diffusion based uni-condition models, and provide the rigorous theory proof. This offers a novel perspective for multi-condition image synthesis task.
\item Extensive experiments demonstrate the flexibility of our proposed framework in integrating multiple diffusion-based T2I generation models, as well as its superior performance on controllable and high-fidelity image generation. 
\end{itemize}

\section{Related Work}

\vspace{-0.4em}
\subsection{Controllable Image Synthesis}
Layout, as a versatile and user-friendly conditional input, has been extensively explored in the field of controllable image generation. Layouts are typically categorized into bounding boxes, segmentation masks \cite{ALDM,SceneComposer}, scribbles \cite{Scribble,ControlNet}, etc. InstanceDiffusion \cite{InstanceDiffusion} has achieved notable success in layout-to-image synthesis by converting various layouts into point representations, allowing a single model to accept different types of condition inputs. However, at its core, it still only processes point-based conditions.

Some approaches transform layouts into textual representations and concatenate them with textual descriptions \cite{Reco}, but incorporating different layout combinations or adding additional conditions still necessitates retraining the model, which is both time-consuming and costly. In certain cases, finer control is required, such as specifying a person’s pose through keypoints for hands or body \cite{handiffuser,gligen}, or even 3D meshes \cite{3d-hand,3d-hand2}. Additionally, image synthesis models often struggle with generating visual text, i.e., signs and banners in the generated images often contain spelling mistakes or incorrectly rendered letters.  To address this, some studies have incorporated text layouts into diffusion models, enabling accurate visual text generation in the image synthesis task \cite{textdiffuser,textdiffuser-2}. 
Moreover, other works have tackled image-guided image generation \cite{plug}, created images in various styles (such as oil painting and cartoon styles) \cite{sgdiff}, and generated images with specific faces \cite{ip}. 

These advancements have greatly enriched the field of controllable T2I synthesis. Unfortunately, each of these models only supports one type of control condition input, heavily hindering their potential application in controllable image generation. Our work, a training-free framework for integrating diffusion-based image generation models, has become increasingly necessary for supporting more flexible and diverse control conditions in image generation.

\begin{figure*}[htbp]
    \centering
    \includegraphics[width=0.8\textwidth]{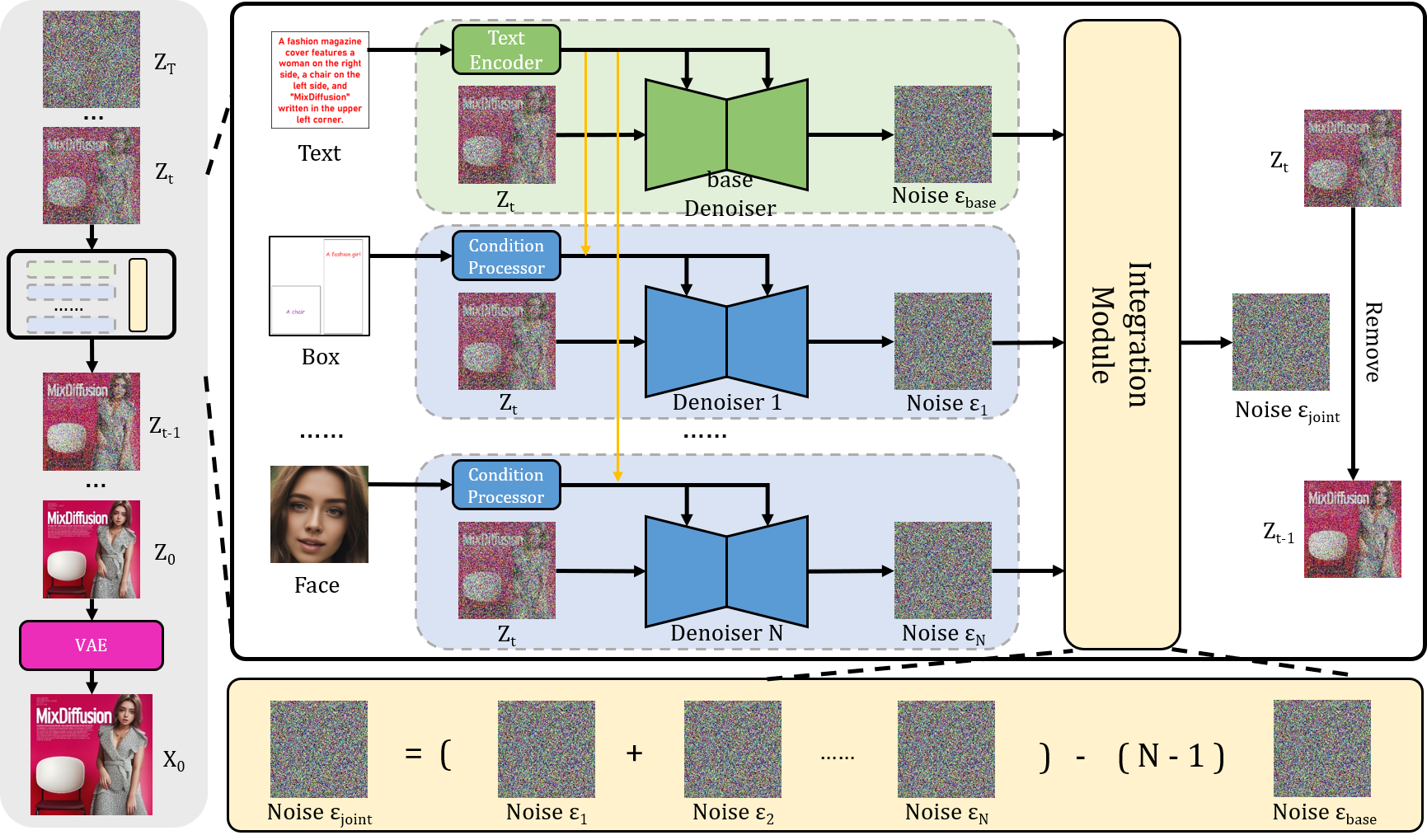}
    \caption{The overall architecture of MixDiffusion, which follows the denoising strategy of diffusion model. In each deonising step, first, each control condition inputs is processed by the corresponding Condition Processor to extract feature. Then, each pre-trained uni-condition denoiser accept condition feature to predict a noise. Finally, the predicted noises from each uni-condition model are integrated by the Integration Module
to get the final predicted noise, which will be removed in this denoising step of MixDiffusion.}
    \label{mixdiffusion}
\end{figure*}

\subsection{Multi-condition Image Synthesis}
The growing interest in controllable image generation has led to the development of models such as Collaborative Diffusion \cite{collaborative-diffusion}, Diffblender \cite{diffblender}, Uni-ControlNet \cite{zhao2024uni} and T2I-Adapter \cite{ip}, which can accept more than one condition inputs to guide the image synthesis. However, these models face limitation due to the difficulty in collecting and annotating diverse training datasets, as well as the high training costs. Furthermore, they often support only two or three fixed control condition inputs, and modifying input combinations or adding new type of control conditions typically requires retraining the model.

In contrast, the proposed MixDiffusion overcomes this limitation by mixing existing pre-trained diffusion-based image generation models to flexibly support various control conditions for controllable image generation in a training-free manner. This allows greater adaptability and efficiency for finer controllable image generation with multiple condition inputs.

\section{Preliminaries: Stable Diffusion}

Stable Diffusion \cite{sd} consists of two primary components:  an autoencoder consisting of an encoder $\mathcal{E}$ and a decoder $\mathcal{D}$, and a denoiser $\epsilon_\theta$. The encoder $\mathcal{E}$ maps images into a latent feature space, i.e., $z_0 = \mathcal{E}(x_0)$ where $z_0$ is the encoded image feature and $x_0$ represents the original image, reducing feature dimensionality to save computational memory and speed up inference. The decoder $\mathcal{D}$ is responsible for mapping the denoised feature back into the image space, i.e., $x_0 = \mathcal{D}(z_0)$. 

The diffusion process is divided into a \textit{forward process} and a \textit{reverse process}. The forward process, which is used in training, incrementally adds small amounts of standard Gaussian noise $\epsilon$ to a clean latent image $z_0$, eventually transforming it into standard Gaussian noise $z_T$. In contrast, in the reverse process, Gaussian noise $z_T$ is sampled and progressively denoised with the denioser $\epsilon_\theta$ over $T$ steps, transforming the noise $z_T$ into a clean latent image $z_0$. At each step, the denoiser $\epsilon_\theta$ predicts the noise $\epsilon_t$ to be removed,
\begin{equation} \epsilon_t = \epsilon_\theta(z_t, t) \end{equation}
\begin{equation} z_{t-1} = \frac{z_t - (1 - \sqrt{1 - \overline{\alpha_t}})\epsilon_t}{\sqrt{\overline{\alpha_t}}}, \label{predict_z_t-1} \end{equation}
where $\lbrace\overline{\alpha_t}\rbrace_{t=1}^T$ are hyperparameters, $t \in T$ represents the current time step, and $\theta$ denotes the parameters of the denoiser model $\epsilon_\theta$. The model $\epsilon_\theta$ can then be trained via:
\begin{equation} \mathcal{L} = \mathbb{E}_{z\sim\mathcal{E}(x),\epsilon\sim\mathcal{N}(0,1),t}[||\epsilon-\epsilon_\theta(z_t,t)||_2^2]. \end{equation}

\vspace{-0.25em}
\section{Method}
We first introduce the framework of the proposed MixDiffusion. Then, the integration formula for MixDiffusion is derived. Finally, we introduce an adjustment strategy of the integration intensity, aiming at further improving the quality of the generated image and the consistency between generated image and control conditions. Furthermore, we have expanded the mathematical derivations of MixDiffusion in different diffusion frameworks, e.g., flow matching.

\subsection{Framework of MixDiffusion} \label{arch}
We divide a controllable image generation model into four key components: a Condition Processor for preprocessing the control conditions, a Text Encoder, a denoiser (i.e., UNet or Transformer), and a Variational Autoencoder (VAE). Usually, each uni-condition image generation model comes with its own pre-trained VAE and Text Encoder. It will be quite memory-consuming and computation-costing to use separate VAE and Text Encoder for each uni-condition model.  Fortunately, we find that when these uni-condition models share the same VAE and Text Encoder, they can maintain strong performance and generate high-quality images. Therefore, the same VAE and Text Encoder are adopted for all these models, which significantly streamlines the integration process. As shown in Fig.~2, the proposed MixDiffusion follows the same denoising strategy of diffusion models. In every denoising step of MixDiffusion, each control condition is first processed by the pre-trained condition processor to obtain the condition feature, and then input into the pre-trained denoiser of the corresponding uni-condition diffusion model to predict a noise. These predicted noises are integrated by the Integration Module to get the final integrated noise, which will be removed in this denoising step of MixDiffusion. After $T$ iterations of denoising, the final latent is obtained. Subsequently, this latent is decoded by the VAE to produce the final generated image. 

\subsection{Integration Module} \label{strategy}
Each denoising step of the reverse diffusion process can be interpreted as sampling a noise from a Gaussian distribution:
\begin{equation}
    \epsilon_{t-1} \sim \mathcal{N}\!\big(u_t(z_t, t), \sigma_t^2 I\big),
\end{equation}
where $u_t$ represents the mean of the Gaussian distribution at step $t$, predicted by the denoiser $\epsilon_\theta$ based on the latent variable $z_t$ and the current step $t$. $\sigma_t$ denotes the standard deviation of the predicted Gaussian distribution. In this work, we employ the DDIM scheduler \cite{DDIM}, where the values $\{\sigma_t\}_{t=1}^T$ are constants, and the denoiser only predicts $u_t$. This simplification enhances the inference process and contributes to a more elegant final integration formula.

The core idea of MixDiffusion's Integration Module is to treat the predicted noises of uni-condition controllable image generation models as distributions
$\{\mathcal{N}(\epsilon_{t-1}; u_{t,i}(z_t, C_i), \sigma_t^2I)\}_{i=1}^N$,
where $C_i$ represents the condition input and $N$ indicates the number of integrated uni-condition models. These distributions are utilized to compute the predicted noise of the integrated model MixDiffusion, which can be treated as a virtually trained multi-condition image generation model accepting multiple condition inputs,
$\mathcal{N}(\epsilon_{t-1}; u_t(z_t, C_1, C_2, \dots, C_N), \sigma_t^2I)$.

We assume a common base (unconditional) distribution $P(\epsilon_{t-1}\mid z_t)$ and conditional independence of conditions given $(\epsilon_{t-1}, z_t)$, i.e.,
\begin{equation}
P(C_1,\dots,C_N \mid \epsilon_{t-1}, z_t) = \prod_{i=1}^N P(C_i \mid \epsilon_{t-1}, z_t).
\end{equation}
Under these assumptions, Bayes' rule yields a product-of-experts style fusion:
\begin{equation}\label{p}
\begin{aligned}
P(\epsilon_{t-1} \mid C_1, \dots, C_N, z_t)
&\propto P(\epsilon_{t-1}\mid z_t)\prod_{i=1}^N P(C_i \mid \epsilon_{t-1}, z_t) \\
&\propto \frac{\prod_{i=1}^N P(\epsilon_{t-1}\mid C_i, z_t)}{P(\epsilon_{t-1}\mid z_t)^{\,N-1}} .
\end{aligned}
\end{equation}
Here, $P(\epsilon_{t-1} \mid C_i, z_t)$ denotes the distribution of the $i$-th uni-condition controllable diffusion model, while $P(\epsilon_{t-1}\mid z_t)$ corresponds to a base diffusion model that does not accept any additional condition input.

We select the sample with the highest probability density (equivalently, the mode/MAP estimate) as the sampling noise of the integrated model, leading to a concise integration formula. With equal variance $\sigma_t^2$ shared across models (DDIM setting),
\begin{equation}
\begin{aligned}
\epsilon_{\mathrm{joint}}
&= \mathop{\arg\max}\limits_{\epsilon}\;
\frac{\prod \limits_{i=1}^N P(\epsilon \mid C_i, z_t)}
     {P(\epsilon \mid z_t)^{\,N-1}} \\
&= \mathop{\arg\max}\limits_{\epsilon}\;
\frac{\prod \limits_{i=1}^N
\frac{1}{\sqrt{2\pi}\sigma_t}\exp\!\left(-\frac{(\epsilon-\epsilon_i)^2}{2\sigma_t^2}\right)}
{\left[\frac{1}{\sqrt{2\pi}\sigma_t}\exp\!\left(-\frac{(\epsilon-\epsilon_{\mathrm{base}})^2}{2\sigma_t^2}\right)\right]^{N-1}} \\
&= \mathop{\arg\min}\limits_{\epsilon}\;
\sum_{i=1}^N(\epsilon-\epsilon_i)^2 - (N-1)(\epsilon-\epsilon_{\mathrm{base}})^2 \\
&= \sum_{i=1}^N \epsilon_i - (N-1)\epsilon_{\mathrm{base}} .
\end{aligned}
\label{integration formula}
\end{equation}
where $\epsilon_{\mathrm{joint}}$ denotes the predicted noise of the integrated multi-condition image generation model MixDiffusion that supports multiple control conditions, $\epsilon_i$ represents the predicted noise of each individual uni-condition controllable diffusion model, and $\epsilon_{\mathrm{base}}$ indicates the predicted noise of a diffusion model that does not accept additional condition input.

\vspace{-0.5em}
\subsection{Integration Intensity Strategy}  \label{intensity}
To further improve the image quality and the consistency between control conditions and generated image content, we adopt a dynamic integration intensity for each denoising step during the $T$ denoising steps of the diffusion model. The early steps primarily determine the object's contours and positions, while the later steps focus on depicting image details \cite{ControlNet, guo2025generateimagescotlets}. Consequently, in our integration strategy, we gradually decrease the integration strength in the denoising steps, allowing the base denoiser to serve as the primary denoiser in the later steps to draw more content details. The mathematical formula for adjusting the integration intensity is as follows:
\begin{equation}
    \epsilon_{t-1} = w\,\epsilon_{\mathrm{joint}} + (1-w)\,\epsilon_{\mathrm{base}},
\end{equation}
where $w$ represents the integration strength, which is modeled as a quadratic function of the steps,
\begin{equation}
    w = (t/T)^2 \in [0,1].
\end{equation}
This integration intensity adjustment strategy ensures that the image details are less susceptible to errors, resulting in a more coherent generated image.

\subsection{Generalization to Flow Matching} 
\label{flow_matching_extension}

The above integration rule is derived under the DDIM formulation, where 
each controllable model predicts the noise variable $\epsilon$. 
The proposed integration strategy can also be extended to flow matching 
models by replacing the noise prediction with velocity prediction.

Unlike diffusion models, flow matching models learn a continuous velocity 
field:
\begin{equation}
    \frac{d x_t}{dt}=v_\theta(x_t,t),
\end{equation}
where $v_\theta(x_t,t)$ represents the predicted velocity. 
For the $i$-th uni-condition model, we denote the predicted velocity as:
\begin{equation}
    v_i=v_\theta(x_t,t,C_i),
\end{equation}
and the unconditional base model predicts:
\begin{equation}
    v_0=v_\theta(x_t,t).
\end{equation}

To consider different prediction uncertainties among models, we model each 
velocity prediction as a Gaussian expert:
\begin{equation}
    P(v_t|C_i,x_t)
    =
    \mathcal{N}(v_i,\sigma_i^2I),
\end{equation}
and the base velocity distribution is defined as:
\begin{equation}
    P(v_t|x_t)
    =
    \mathcal{N}(v_0,\sigma_0^2I).
\end{equation}

Following the same product-of-experts formulation, the integrated velocity 
distribution is given by:
\begin{equation}
\begin{aligned}
P(v_t|C_{1:N},x_t)
&\propto
\frac{
\prod_{i=1}^{N}P(v_t|C_i,x_t)
}
{
P(v_t|x_t)^{N-1}
}.
\end{aligned}
\end{equation}

Substituting the Gaussian distributions and removing constant terms, 
maximizing the above probability is equivalent to minimizing:
\begin{equation}
\begin{aligned}
\mathcal{L}(v_t)
&=
\sum_{i=1}^{N}
\frac{(v_t-v_i)^2}{2\sigma_i^2}
-
(N-1)
\frac{(v_t-v_0)^2}{2\sigma_0^2}.
\end{aligned}
\end{equation}

Taking the derivative with respect to $v_t$ and setting it to zero gives:
\begin{equation}
\begin{aligned}
&
\sum_{i=1}^{N}
\frac{v_t-v_i}{\sigma_i^2}
-
(N-1)
\frac{v_t-v_0}{\sigma_0^2}
=0 .
\end{aligned}
\end{equation}

Rearranging the terms associated with $v_t$, we obtain:
\begin{equation}
\begin{aligned}
v_t
\left(
\sum_{i=1}^{N}\frac{1}{\sigma_i^2}
-
(N-1)\frac{1}{\sigma_0^2}
\right)
&=
\sum_{i=1}^{N}
\frac{v_i}{\sigma_i^2}
-
(N-1)
\frac{v_0}{\sigma_0^2}.
\end{aligned}
\end{equation}

Therefore, the fused velocity field is:
\begin{equation}
\boxed{
v_{\mathrm{joint}}
=
\frac{
\displaystyle
\sum_{i=1}^{N}
\frac{v_i}{\sigma_i^2}
-
(N-1)
\frac{v_0}{\sigma_0^2}
}
{
\displaystyle
\sum_{i=1}^{N}
\frac{1}{\sigma_i^2}
-
(N-1)
\frac{1}{\sigma_0^2}
}
}.
\label{fm unequal variance}
\end{equation}

The above formulation provides an uncertainty-aware velocity fusion 
strategy, where experts with smaller variance contribute more strongly to 
the final velocity prediction. In the special case where all experts share 
the same variance, i.e.,
$\sigma_i^2=\sigma_0^2$, Eq.~\eqref{fm unequal variance} reduces to:
\begin{equation}
    v_{\mathrm{joint}}
    =
    \sum_{i=1}^{N}v_i
    -(N-1)v_0 .
\end{equation}

Correspondingly, for DDIM diffusion models with different prediction 
variances, the same derivation can be applied by replacing the velocity 
field with the predicted noise. The resulting noise fusion rule is:
\begin{equation}
\epsilon_{\mathrm{joint}}
=
\frac{
\displaystyle
\sum_{i=1}^{N}
\frac{\epsilon_i}{\sigma_i^2}
-
(N-1)
\frac{\epsilon_0}{\sigma_0^2}
}
{
\displaystyle
\sum_{i=1}^{N}
\frac{1}{\sigma_i^2}
-
(N-1)
\frac{1}{\sigma_0^2}
}.
\label{ddim unequal variance}
\end{equation}

When the DDIM scheduler uses a shared variance for all models, 
Eq.~\eqref{ddim unequal variance} degenerates to the original integration 
rule:
\begin{equation}
    \epsilon_{\mathrm{joint}}
    =
    \sum_{i=1}^{N}\epsilon_i
    -(N-1)\epsilon_0 .
\end{equation}

\section{Experiment}
In this section, we outline the experimental setup, including models, hyper-parameter settings, datasets, and evaluation metrics. And then, the proposed MixDiffusion is compared with state-of-the-art multi-condition image generation models. Then ablation study is conducted to analyze the contribution of each proposed module/strategy to MixDiffusion. Finally, we present quantitative results that compare the consistency of generated images with text descriptions and assess image quality after integrating various numbers of models. 

\subsection{Experiments Setup} \label{setup}
Theoretically, the proposed MixDiffusion can mix unlimited diffusion-based uni-condition T2I generation models, which means that MixDiffusion can take unlimited control conditions as input to control the image generation, as long as there are pre-trained diffusion-based uni-condition image generation models supporting these control conditions. In this work, we use six pre-trained diffusion-based image generation models to illustrate the performance of MixDiffusion, TextDiffuser V2 \cite{textdiffuser-2}, T2I-Adapter \cite{t2i}, Gligen \cite{gligen}, IP-Adapter \cite{ip}, DreamShaper \cite{lykon2023dreamshaper}, and Stable Diffusion V1.5 \cite{sd}, which support different control conditions beyond textual descriptions, including bounding boxes, poses, faces, text layouts, segmentation maps, etc. Text encoder and VAE are shared by these six models to reduce computational memory and time. 
All experiments were conducted on a single NVIDIA A800 GPU. The peak GPU memory usage reached 16 GB, and the image generation time was approximately 1.5 times that of Stable Diffusion V1.5 when we run all the denoisers in parallel.

Due to its training-free nature, the proposed MixDiffusion does not require a training stage. We directly evaluate its performance using subsets of the COCO2017Eval \cite{lin2014microsoft} dataset which contains images of people. To assess consistency with human preferences and the realism of generated images, we adopt the ImageReward score \cite{ImageReward}. A higher ImageReward score indicates better alignment with human preferences. We also use a pretrained model \cite{aesthetic-predictor-v2-5} to evaluate the aesthetic quality of the generated images.

To evaluate how well the generated images adhere to control conditions, we use various metrics: for bounding boxes, we compute the Intersection over Union (IoU) between the predicted and ground-truth boxes; for keypoints, we apply Object Keypoint Similarity (OKS) between the predicted and ground-truth input keypoints. For structural conditions such as canny edges and depth maps, we use the Root Mean Square Error (RMSE) to quantify pixel-level deviations between the generated results and the control inputs—lower RMSE values indicate better preservation of edge details and depth information.

\begin{table}
  \centering
  \resizebox{1.\linewidth}{!}{%
      \begin{tabular}{lccc}
        \toprule
        Methods & Depth(RMSE)$\downarrow$ & Canny(RMSE)$\downarrow$ & OKS(AP@0.5)$\uparrow$ \\ 
        \midrule
        CnC & 35.42 & - & - \\
        Cocktail & 32.78 & - & \underline{75.82\%} \\
        T2I-Adapter & \textbf{30.51} & \underline{91.78} & - \\
        Uni-ControlNet & 31.75 & 93.21 & 67.67\% \\
        Diffblender & 33.29 & 102.21 & 48.37\% \\
        AnyControl & 32.97 & 105.65 & 59.50\% \\
        DynamicControl & 33.92 & 103.89 & 76.55 \\
        \rowcolor{gray!20}
        MixDiffusion & \underline{31.21} & \textbf{90.32} & \textbf{86.03\%} \\
        \bottomrule
      \end{tabular}
  }
  \caption{Comparison of multi-control methods on COCO2017 with four control conditions (Text, Depth, Canny and Pose).}
  \label{cmp2}
\vspace{-2em}
\end{table}

\subsection{Comparison to Previous Methods} \label{comparison}
Since there is no training-free multi-condition image generation model, we compare the proposed MixDiffusion with several state-of-the-art methods for multi-condition image generation, including Uni-ControlNet \cite{zhao2024uni}, Diffblender \cite{diffblender}, AnyControl \cite{anycontrol}, Cocktail \cite{hu2023cocktail}, DynamicControl \cite{he2025dynamiccontroladaptiveconditionselection}, and CnC \cite{lee2024compose}, all of which require training with different combinations of control inputs.

As shown in Table \ref{cmp2}, our proposed MixDiffusion outperforms other models in control accuracy. This demonstrates that MixDiffusion can generate images that follow the control conditions much better than the other models, regardless of the combinations of control inputs.
This is because, compared to fixed models that require training, MixDiffusion can flexibly combine existing strong-performing uni-condition models. Additionally, the multiple condition inputs can mutually verify and supplement each other, thus further enhancing the performance of these already effective uni-condition models.

\begin{table}[htbp]
  \centering
  \small  
  \begin{tabular}{@{}c@{\hspace{0.5em}}cc@{\hspace{0.5em}}c@{\hspace{0.5em}}c@{}}
    \toprule
    \multirow{2}{*}{Settings} & \multicolumn{2}{c}{Ablation Settings} & \multicolumn{2}{c}{IoU} \\
    \cmidrule(lr){2-3} \cmidrule(lr){4-5}
    & Interpolation & No Intensity & AP@0.5$\uparrow$ & AP@0.75$\uparrow$ \\
    \midrule
    \multirow{3}{*}{TBD}   & $\CheckedBox$ & $\square$ & 53.96\% & 38.03\% \\
                           & $\square$ & $\CheckedBox$ & 67.17\% & 56.27\% \\
                           & $\square$ & $\square$ & \textbf{86.57\%} & \textbf{72.00\%} \\
    \midrule
    \multirow{3}{*}{TBDS}  & $\CheckedBox$ & $\square$ & 59.27\% & 45.45\% \\
                           & $\square$ & $\CheckedBox$ & 66.44\% & 55.56\% \\
                           & $\square$ & $\square$ & \textbf{87.79\%} & \textbf{72.57\%} \\
    \midrule
    \multirow{3}{*}{TBDSC} & $\CheckedBox$ & $\square$ & 63.02\% & 50.90\% \\
                           & $\square$ & $\CheckedBox$ & 65.42\% & 55.28\% \\
                           & $\square$ & $\square$ & \textbf{88.51\%} & \textbf{73.36\%} \\
    \bottomrule
  \end{tabular}
  \caption{Ablation study on the COCO 2017 dataset. A checkmark in the "Interpolation" column denotes integration strategy was replaced with interpolation , while a checkmark in the "No Intensity" column denotes the removal of the Integration Intensity module from MixDiffusion. The condition inputs are (T)ext, (B)ox, (D)epth, (S)ketch, and (C)anny.}
  \label{ab:bbox}
\vspace{-1.3em}
\end{table}

\subsection{Ablation Study} \label{ablation}
In this section, we replace MixDiffusion's integration strategy with interpolation and remove the integration intensity to respectively evaluate the contributions of these two modules to the overall performance of MixDiffusion.

\begin{figure*}[t]
    \centering
    \includegraphics[width=0.8\textwidth]{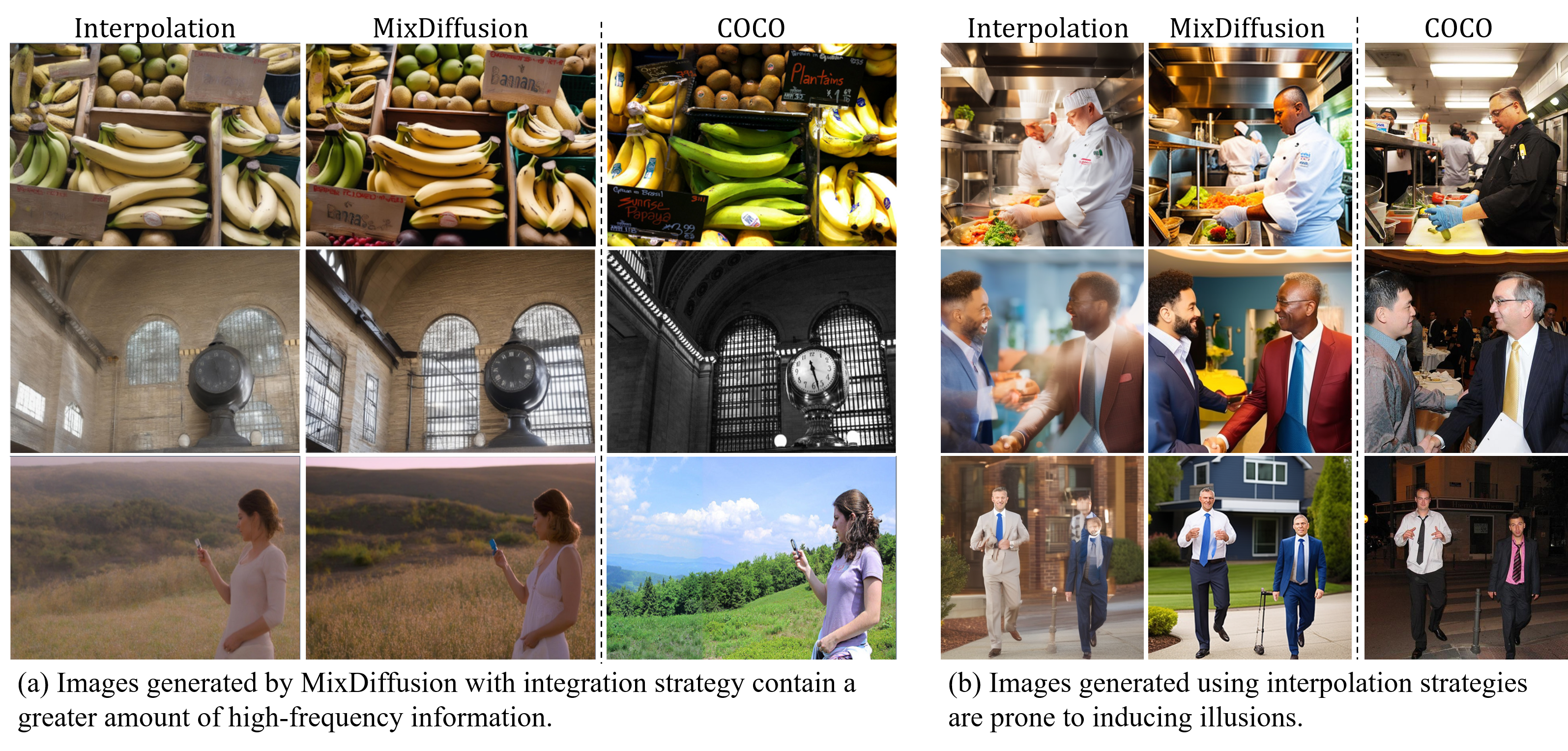}
    \caption{Comparison of images generated using the interpolation and the integration strategy. Experiments conducted on the COCO2017 Evaluation Set. 'COCO' stands for original reference images in the dataset.}
    \label{hf}
\vspace{-1.0em}
\end{figure*}

\subsubsection{Integration Strategy}
Instead of using the proposed integration strategy to mix uni-condition image generation models in MixDiffusion, a more straightforward way is to interpolate these models, i.e., taking the average of the predicted noises of each uni-condition models as the final predicted noise,
\begin{equation} \epsilon_{t-1} = \frac{\sum^N_{i=1}{\epsilon_i}}{N}. \end{equation}

To demonstrate the effectiveness of the proposed integration strategy, we integrate five uni-condition diffusion models with the proposed integration strategy and the interpolation strategy, respectively. The five models receive conditional inputs of text (T), bounding boxes (B), canny edge (C), depth (D), and sketch (S) respectively. The quantitative comparison results are shown in Table \ref{ab:bbox}, which indicates that when integrating a larger number of models, the interpolation method fails to align the generated images closely with control conditions, as shown in the first row of TBD, TBDS and TBDSC. This can be explained through the interpolation formula, which suggests that as more models are integrated, the contribution of each component diminishes, weakening the overall control strength. In contrast, the integration strategy Eq.~(\ref{integration formula}) does not encounter this issue, as the contribution of each control model remains constant regardless of the number of models. Moreover, the images generated by the interpolation method are with lower quality, as depicted in Fig.~\ref{hf}, which loses significant high-frequency information and introduce faint artifacts due to the direct summation of predictions from different models.

\subsubsection{Integration Intensity}
Table \ref{ab:bbox}  also compares the image generation results of MixDiffusion with and without the integration intensity adjustment strategy (i.e., the second row V.S. the third row in TBD, TBDS and TBDSC) in terms of IoU. The results show that the model consistently performs better when intensity control is applied.  MixDiffusion with integration intensity adjustment produces images with a more harmonious integration of components and better alignment with the conditional inputs compared to MixDiffusion without integration intensity adjustment. This improvement stems from an important phenomenon: in the later denoising steps of generation, the predictions of individual models diverge, and their combination can introduce competition, resulting in less cohesive outputs.

\begin{table}[htbp]
  \centering
  \small  
  \setlength{\tabcolsep}{3pt}  
  \renewcommand{\arraystretch}{0.8}
  
  \begin{tabular}{@{}lccccc@{}}
    \toprule
    \multirow{2}{*}{Settings} & Aesthetic & \multicolumn{2}{c}{ImageReward} & \multicolumn{2}{c}{IoU} \\ 
    \cmidrule(lr){3-4} \cmidrule(lr){5-6}
    & Score$\uparrow$ & rating$\uparrow$ & artifact$\downarrow$ & AP@0.5 $\uparrow$ & AP@0.75 $\uparrow$ \\ 
    \midrule
    TB       & 4.46  & 4.30 & 2.92 & 84.55\% & 69.86\% \\
    TBD      & 4.78 & 4.47 & 2.88 & 86.57\% & 72.00\% \\
    TBDS     & \underline{4.92} & \textbf{4.73} & \underline{2.82} & \underline{87.79\%} & \underline{72.57\%} \\
    TBDSC    & \textbf{5.15} & \textbf{4.73} & \textbf{2.80} & \textbf{88.51\%} & \textbf{73.36\%} \\
    \midrule[\heavyrulewidth]
  \end{tabular}
  
  \begin{tabular}{@{}lcccccc@{}}
    \toprule
    \multirow{2}{*}{Settings} & Aesthetic & \multicolumn{2}{c}{ImageReward} & \multicolumn{2}{c}{OKS} \\ 
    \cmidrule(lr){3-4} \cmidrule(lr){5-6}
    & Score $\uparrow$ & rating$\uparrow$ & artifact$\downarrow$ & AP@0.5 $\uparrow$ & AP@0.75 $\uparrow$ \\ 
    \midrule
    TP       & 4.70 & 3.73 & 3.24 & 85.10\% & 45.94\% \\
    TPD      & 4.75 & 3.97 & 3.19 & 85.98\% & 46.96\% \\
    TPDS     & \underline{4.97} & \underline{4.15} & \underline{3.14} & \underline{86.03\%} & \textbf{47.87\%} \\
    TPDSC    & \textbf{5.33} & \textbf{4.19} & \textbf{3.13} & \textbf{88.09\%} & \underline{47.30\%} \\
    \bottomrule
  \end{tabular}
  
  \caption{Evaluation results on COCO2017 Val. Settings: T(ext)+B(Box)+D(epth)+S(ketch)+C(anny) for bounding box conditions; T(ext)+P(ose)+D+S+C for pose conditions. When the settings are TB and TP, they correspond to the individual uni-control models.}
  \label{tab:combined_evaluation}
\vspace{-1.3em}
\end{table}

\subsection{Further Analysis of the Integration Module} \label{qulitative}
To further demonstrate the effectiveness of the proposed integration strategy, we conduct experiments to generate images under two conditions (text and bounding box (TB), also text and pose (TP)), three conditions (adding depth D), four conditions (adding sketch S), and five conditions (adding canny edge C). As shown in Table \ref{tab:combined_evaluation}, the OKS, IoU, aesthetic scores and ImageReward improve with increasing control conditions. 
This is because, as the number of condition inputs increases, more details are exposed to the model. These inputs complement and verify one another, resulting in outputs that better align with expectations. Particularly when inputs such as canny edges or sketches are used, they provide the model with substantial detail about the foreground and background, which in turn enhances the model’s adherence to the other given conditions.

\section{Conclusion and Future Work}

In this work, we propose MixDiffusion, a training-free controllable image generation framework that integrates multiple off-the-shelf uni-condition diffusion models for flexible multi-condition synthesis. By deriving a principled integration formula from a product-of-experts perspective, MixDiffusion provides an effective solution for combining different control signals without additional training. Extensive experiments demonstrate its superiority in both image quality and condition alignment. As shown in Fig.~\ref{headfigure} and the supplementary MixDiffusion Gallery, MixDiffusion is particularly effective when combining complementary conditions from different aspects, such as foreground, background, and style.

Despite its effectiveness, MixDiffusion has several limitations. The current formulation relies on a conditional independence assumption among different controls. While this assumption generally works well for complementary conditions, conflicting conditions may lead to ambiguous results due to the competition between constraints. Future work will explore more effective strategies for handling condition dependencies and conflicts. Moreover, although MixDiffusion can be extended to Flow Matching models, recent advances in Flow Matching-based model merging and LoRA combination suggest promising potential for further exploration. Finally, since multiple denoisers are involved during inference, MixDiffusion may require additional computational resources and memory, even though parallel execution can alleviate the increase in inference time.

\bibliography{aaai2027}

\clearpage
\appendix

\section{Appendix}
\subsection{Proof of the Integration Formula} \label{A}
In this section, we provide a detailed derivation of the integration formula. We first establish the relationship between the Gaussian distributions of the uni-condition image generation models and the one of the multi-condition image generation model. Following that, we derive the integrated noise for the multi-condition image generation model, which is removed in each denoising step,  from the individual noises of the uni-condition image generation models.

\begin{equation}
    \begin{aligned}
       P(\epsilon_{t-1} |& C_1, C_2, \dots C_N, z_t)   \\
       & = \frac{P( C_1, C_2, \dots C_N, \epsilon_{t-1} | z_t)}{P(C_1, C_2, \dots C_N | z_t)} \\
       & = \frac{P( C_1, C_2, \dots C_N | \epsilon_{t-1}, z_t)P(\epsilon_{t-1}|z_t)}{P(C_1, C_2, \dots C_N | z_t)} \\
       & \overset{\text{\textcircled{1}}}{=} \frac{(\prod \limits_{i=1}^N P( C_i | \epsilon_{t-1}, z_t))P(\epsilon_{t-1}|z_t)}{\prod \limits_{i=1}^N P(C_i | z_t)} \\
       & = \frac{(\prod \limits_{i=1}^N \frac{P(\epsilon_{t-1} | C_i, z_t) P(C_i | z_t)}{P(\epsilon_{t-1} | z_t)} )P(\epsilon_{t-1}|z_t)}{\prod \limits_{i=1}^N P(C_i | z_t)} \\
       & = \frac{\prod \limits_{i=1}^N P(\epsilon_{t-1} | C_i, z_t)}{\prod \limits_{i=1}^{N-1} P(\epsilon_{t-1} | z_t)}
    \end{aligned}
    \label{proofeq5}
\end{equation}

In \textcircled{1}, we assume that the classifiers of each model are independent of each other, i.e., $P( C_1, C_2, \dots C_N | \epsilon_{t-1}, z_t)=\prod \limits_{i=1}^N P(C_i | \epsilon_{t-1}, z_t)$. Intuitively, this assumption aligns with human cognition, since each diffusion-based uni-condition model is trained independently.

To determine the sampling result of the multi-condition image generation model, we select the sample with the highest probability density. This corresponds to identifying the noise $\epsilon$ that maximizes the conditional probability $P(\epsilon_{t-1} | C_1, C_2, \dots C_N, z_t)$. In other words, we find the noise to be removed by maximizing the Eq.(\ref{proofeq5}).

\begin{equation}
    \begin{aligned}
        \epsilon_{fit}  = &\mathop{\arg\max}\limits_{\epsilon} \frac{\prod \limits_{i=1}^N P(\epsilon | C_i, z_t)}{\prod \limits_{i=1}^{N-1} P(\epsilon | z_t)} \\
        \overset{\text{\textcircled{2}}}{=} &\mathop{\arg\max}\limits_{\epsilon} \frac{\prod \limits_{i=1}^N \frac{1}{\sqrt{2\pi}\sigma} \exp{(-\frac{{(\epsilon-\epsilon_i)}^2}{2\sigma^2})}}{\prod \limits_{i=1}^{N-1} \frac{1}{\sqrt{2\pi}\sigma} \exp{(-\frac{{(\epsilon-\epsilon_{base})}^2}{2\sigma^2})}}    \\
        = &\mathop{\arg\max}\limits_{\epsilon} \frac{1}{\sqrt{2\pi}\sigma} \exp{(-\frac{(\sum_{i=1}^N(\epsilon-\epsilon_i)^2) - (N-1)(\epsilon-\epsilon_{base})^2}{2\sigma^2})}    \\
        = &\mathop{\arg\min}\limits_{\epsilon} \sum_{i=1}^N(\epsilon-\epsilon_i)^2 - (N-1)(\epsilon-\epsilon_{base})^2   \\
        = &\sum_{i=1}^N \epsilon_i - (N-1)\epsilon_{base}
    \end{aligned}
\end{equation}

In \textcircled{2}, we set the standard deviation $\sigma$ of the Gaussian distributions to a constant value. This is because, in DDPM, $\sigma$ could serves as a hyperparameter, and the model only predicts the mean $u$. In other noise schedulers, if the standard deviation varies, the final integration formula will include a term involving $\sigma$ within $\epsilon_i$.

\subsection{Experiment Settings} \label{C}
To make the project reproducable, we introduce the experiment settings in more details.

When calculating COCO OKS, the standard error K for each keypoint is treated as an empirical constants, with our settings as follows:
K = [0.26, 0.25, 0.25, 0.35, 0.35, 0.79, 0.79, 0.72, 0.72, 0.62, 0.62, 1.07, 1.07, 0.87, 0.87, 0.89, 0.89] / 10.0

\begin{figure*}[!h]
    \centering
    \includegraphics[height=0.85\textheight ]{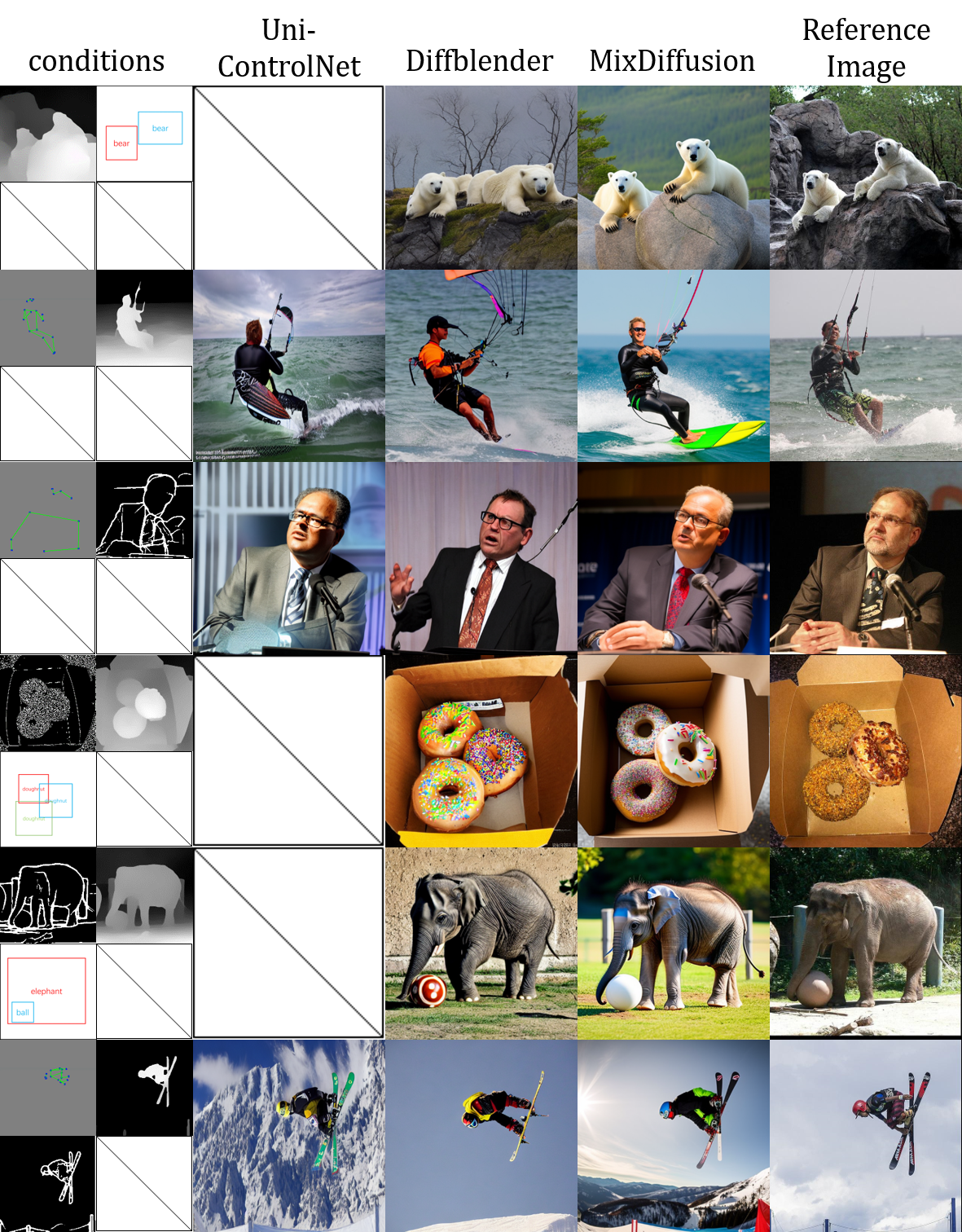}
    \caption{Comparison results of different controllable image generation models. Generation results of the Uni-ControlNet are not presented in the case of bounding box is included in the control condition inputs (i.e., the first rows) since the Uni-ControlNet does not support bounding box input. Uni-ControlNet and DiffBlender perform worse than MixDiffusion in terms of background generation, consistency between generated images and conditions, and image detail quality. }
    \label{a}
\end{figure*}

It is worth noting that Uni-ControlNet does not support bounding box condition inputs. Therefore, in the comparison experiments in which bounding box is included as control condition input, the generation results of the Uni-ControlNet is not presented (e.g., Fig.\ref{a} in this supplementary material). For the individual Uni-condition image models used in the MixDiffusion, we use the recommended hyperparameters and configurations provided in their papers or projects. 

\onecolumn
\subsection{Additional Comparison Result} \label{D}

\begin{figure*}[h]
    \centering
    \includegraphics[height=0.85\textheight ]{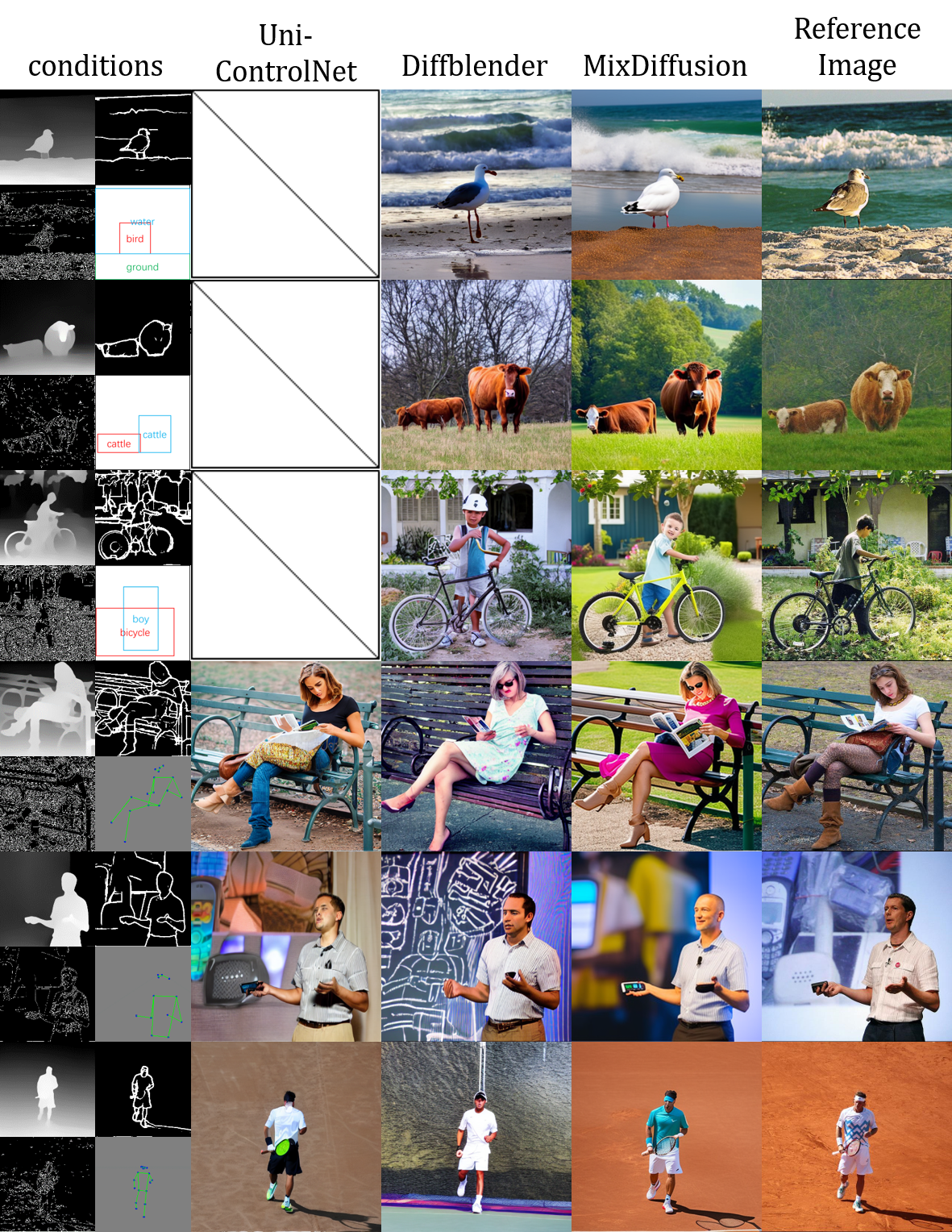}
    \caption{Comparison results of different controllable image generation models. Generation results of the Uni-ControlNet are not presented in the case of bounding box is included in the control condition inputs (i.e., the first three rows) since the Uni-ControlNet does not support bounding box input. Uni-ControlNet and DiffBlender perform worse than MixDiffusion in terms of background generation, consistency between generated images and conditions, and image detail quality. }
    \label{b}
\end{figure*}

\clearpage
\subsection{MixDiffusion Gallery}

\begin{figure*}[!h]
    \centering
    \includegraphics[height=0.85\textheight ]{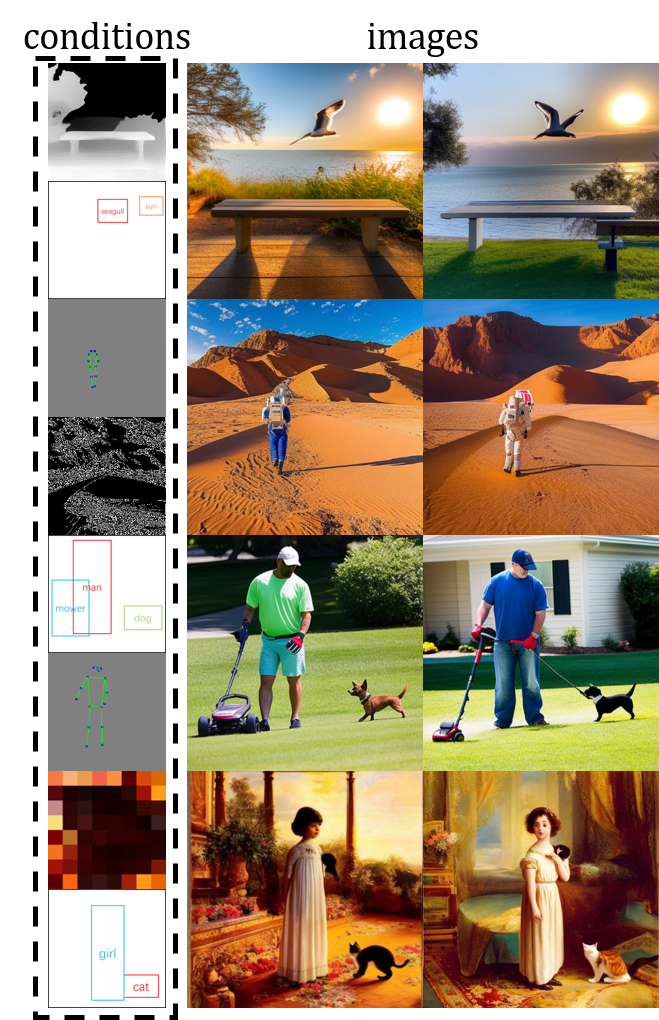}
    \caption{Images generated with our proposed MixDiffusion.}
\end{figure*}

\begin{figure*}[!h]
    \centering
    \includegraphics[height=0.9\textheight ]{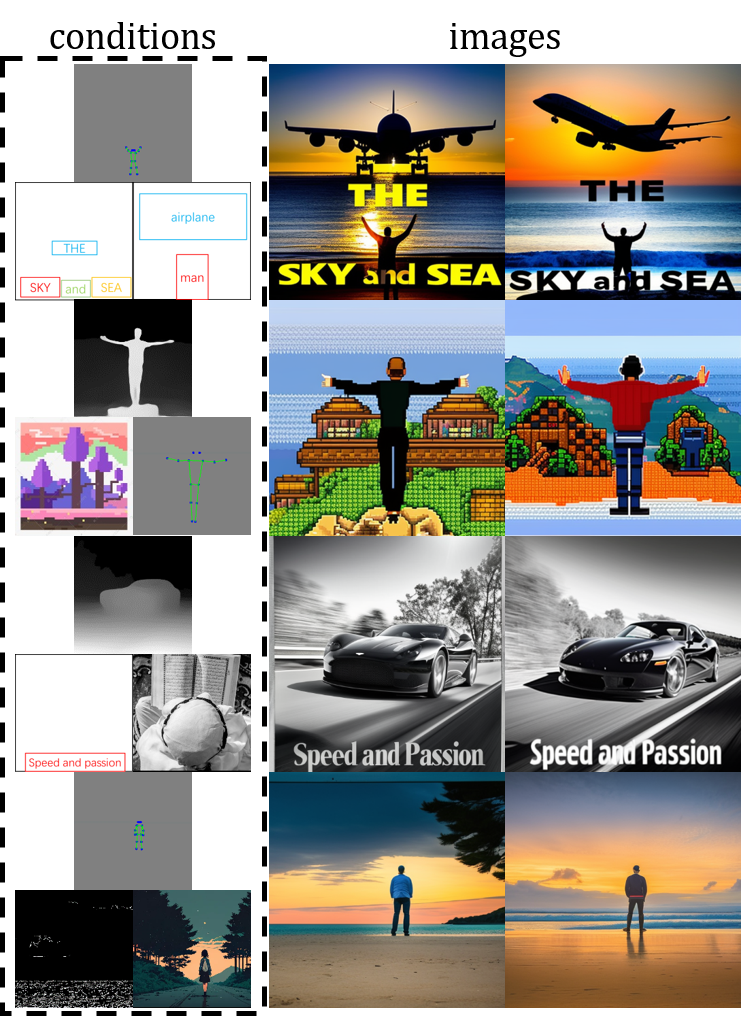}
    \caption{Images generated with our proposed MixDiffusion.}
\end{figure*}

\begin{figure*}[!h]
    \centering
    \includegraphics[height=0.9\textheight]{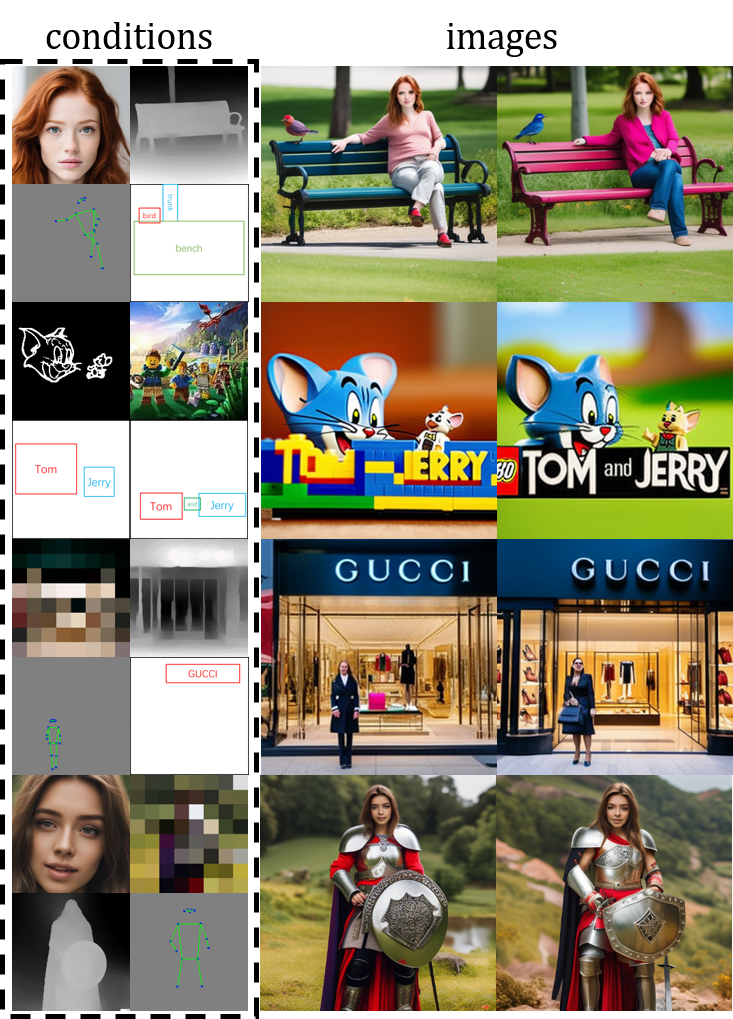}
    \caption{Images generated with our proposed MixDiffusion.}
\end{figure*}

\end{document}